\documentclass[conference]{IEEEtran}
\IEEEoverridecommandlockouts
\usepackage{hyperref}
\usepackage{cite}
\usepackage{amsmath,amssymb,amsfonts}
\usepackage{algorithmic}
\usepackage{graphicx}
\usepackage{textcomp}
\usepackage{cleveref}
\usepackage{xcolor}
\usepackage{xspace}
 \usepackage{multirow} 
\def\BibTeX{{\rm B\kern-.05em{\sc i\kern-.025em b}\kern-.08em
    T\kern-.1667em\lower.7ex\hbox{E}\kern-.125emX}}
\begin{document}
\newcommand{\ie}{\textit{i}.\textit{e}., }
\newcommand{\R}{\mathbb{R}}
\newcommand{\yolovnine}{YOLOv9\xspace}
\newcommand{\regmax}{N}
\newcommand{\moe}{MoE\xspace} 

\title{YOLO Meets Mixture-of-Experts: Adaptive Expert Routing for Robust Object Detection
\\
\thanks{*These authors contributed equally to this work.}
}

\author{
\begin{minipage}[t]{0.36\textwidth}
\centering
Ori Meiraz*\\
\textit{Faculty of Data and Decision Sciences}\\
\textit{Technion, Israel Institute of Technology}\\
Haifa, Israel\\
Email: ori.meiraz@campus.technion.ac.il\\
ORCID: 0009-0001-1469-5582
\end{minipage}
\hfill
\begin{minipage}[t]{0.28\textwidth}
\centering
Sharon Shalev\\
\textit{Independent Researcher}\\
Email: shalev1310@gmail.com \\
ORCID: 0009-0003-5054-230X
\end{minipage}
\hfill
\begin{minipage}[t]{0.36\textwidth}
\centering
Avishai Weizman*\\
\textit{School of Electrical and Computer Engineering}\\
\textit{Ben-Gurion University of the Negev}\\
Beersheba, Israel\\
Email: wavishay@post.bgu.ac.il\\
ORCID: 0009-0004-1182-8601
\end{minipage}
}

\maketitle

\begin{abstract}
This paper presents a novel Mixture-of-Experts framework for object detection, incorporating adaptive routing among multiple \yolovnine-T experts to enable dynamic feature specialization and achieve higher mean Average Precision (mAP) and Average Recall (AR) compared to a single \yolovnine-T model.
\end{abstract}

\begin{IEEEkeywords}
Deep Learning, Object Detection, Mixture-of-Experts, \yolovnine.
\end{IEEEkeywords}

\section{Introduction}
Object detection is one of the most fundamental and challenging problems in computer vision. It aims to identify and localize instances of objects within images or video frames and has received great attention in recent years~\cite{zou2023object,zhao2019object,zou2019review}.

The introduction of \textit{convolutional neural networks} (CNNs) revolutionized the field, giving rise to detectors such as R-CNN \cite{RCNN}, \textit{single-shot detector} (SSD) \cite{liu2016ssd}, and YOLO \cite{wang2024yolov9,jiang2025odverse33newyoloversion,YoloOG}, which achieved remarkable improvements in both accuracy and speed.
Among them, \yolovnine has shown superior performance compared to other models in the YOLO series~\cite{jiang2025odverse33newyoloversion}.

In parallel with these developments, the \textit{Mixture-of-Experts} (MoE)~\cite{mu2025comprehensivesurveymixtureofexpertsalgorithms} paradigm introduces multiple specialized experts and a routing network that dynamically assigns responsibility for each input. The outputs are combined either softly, as weighted mixtures~\cite{shazeer2017}, or through hard expert selection~\cite{switchTransformers}. In~\cite{YoloMoE}, a \moe module is integrated into the YOLOv5~\cite{yolov5} backbone to process intermediate feature maps. The \moe framework has also proven effective in tasks such as next-token prediction~\cite{switchTransformers} and deepfake detection~\cite{negroni2025leveraging}. 
In many of these applications, the outputs correspond to class probabilities or logits, which can be combined straightforwardly as:
\begin{equation}
    \label{eq:straighforward}
z = \sum_{e=1}^{E} \alpha_e z_e
\end{equation}
where $z_e$ denotes the logits produced by expert $e$, and $\alpha_e$ represents the weight assigned by the router, $E$ is the total number of experts, and $z$ is the fused output logits.

Building on these advances, this paper introduces an \moe architecture based on \yolovnine-T (the tiny version of \yolovnine)~\cite{wang2024yolov9}, designed to improve detection performance and robustness.

\begin{figure*}[t] 
    \centering
\includegraphics[width=0.77\linewidth]{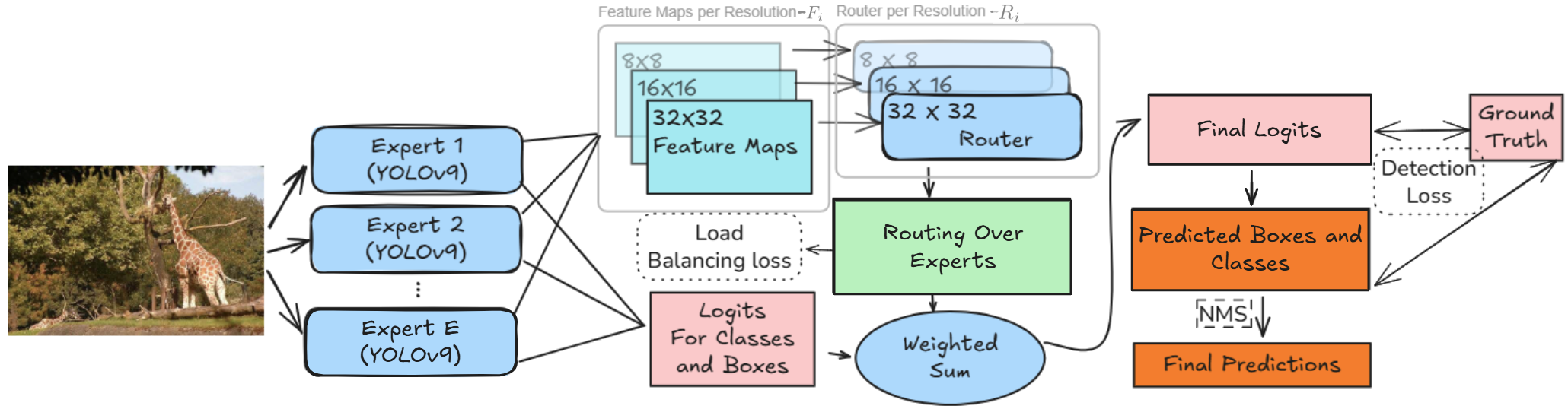}  
    \caption{Visualization of \moe within the \yolovnine architecture, multiple experts process the input image to produce multi-scale feature maps and outputs (class and bounding box logits).
    Routers at different resolutions (8×8, 16×16, 32×32) generate adaptive routing weights that fuse expert outputs into final detections.
    The loss is computed between model outputs before \textit{Non-maximum suppression} (NMS)~\cite{Hosang2017LearningNS,wang2024yolov9} and the ground truth, ensuring end-to-end differentiability.} 
        \label{fig:method} 
\end{figure*}

\section{Proposed MoE-Based \yolovnine Architecture}
In the \yolovnine framework, each input image is processed through $I=3$ feature maps $F_i \in \mathbb{R}^{B \times C_i \times H_i \times W_i}$ corresponding to spatial resolutions $i \in \{8,16,32\}$, where $B$ denotes the batch size, $C_i$ the number of channels, and $(H_i, W_i)$ the spatial dimensions of the $i$-th  level.
Each map is responsible for detecting objects at a corresponding scale, and instead of directly regressing the bounding box coordinates, the model predicts the distances from each cell center to the four sides of the bounding box (left, top, bottom, right).
These distances are discretized into $\regmax$ bins (typically 16) and treated as classification targets, with coordinates obtained by the expectation over predicted probabilities. 
This discrete formulation enables consistent probabilistic fusion across experts (\yolovnine models).

In this framework, each feature level produces a prediction map containing both bounding box and class logits. Specifically, the map outputs 
$ y_{\text{box}} \in \mathbb{R}^{4 \times \regmax} $
for bounding box regression and 
$ y_{\text{cls}} \in \mathbb{R}^{n_c} $
for object classification, where $\regmax$ denotes the number of discrete distance classes used in bounding box regression, and $n_c$ denotes the number of object classes.

Our design follows the Enhanced \moe method proposed in~\cite{negroni2025leveraging}, where the routing decision is based on features extracted by individual experts rather than on the raw input image.
Building on this multi-scale structure of \yolovnine, the resulting feature representations at each level serve as inputs to the \moe routing mechanism.

Each expert $e$ processes its corresponding feature map $F_{i,e} \in \mathbb{R}^{B \times h \times H_i \times W_i}$ and extracts intermediate features from a layer preceding the output head, where $h$ denotes the hidden dimension.
The router $R_i$ at level $i$ operates at its corresponding resolution and receives as input both the outputs of all experts at that level and their concatenated representations, combined through a reweighted Hadamard fusion.
The Hadamard fusion captures cross-expert interactions and is defined as:

\begin{equation}
\label{eq:hadamard_fusion}
K_i = (F_{i,1} \odot F_{i,2} \odot \cdots \odot F_{i,E}) \odot W_i
\end{equation}
where $\odot$ denotes element-wise multiplication, 
and $W_i \in \mathbb{R}^{h \times 1 \times 1}$ is a learnable weighting parameter. 
The combined input to the router is defined as:
\begin{equation}
M_i = \operatorname{Concat}\big[F_{i,1}, F_{i,2}, \ldots, F_{i,E}, K_i\big]
\end{equation}

where $M_i \in \mathbb{R}^{B \times (E+1)h \times H_i \times W_i}$ represents the fused feature map at level $i$, obtained by concatenating all inputs along the channel dimension.
The resulting fused map $M_i$ is processed by the router $R_i$, implemented as a lightweight CNN with downsampling followed by a fully connected layer, to produce $\alpha_i \in \mathbb{R}^{B \times E}$, representing the routing weights assigned to each expert.
A softmax function is applied along the expert dimension to produce normalized routing weights at each $i$ level, ensuring that $\sum_{e=1}^{E} \alpha_{i,e} = 1$.
These weights adaptively modulate the outputs of the experts for both classification and bounding box prediction, as defined in~\autoref{eq:straighforward}.

To prevent the router from collapsing onto a single expert, a load balancing loss inspired by~\cite{switchTransformers} is incorporated. The total loss is defined as:

\begin{equation}
\mathcal{L} = \mathcal{L}_{\text{det}} + \lambda_{\text{lb}}\mathcal{L}_{\text{lb}},
\end{equation}
where $\mathcal{L}_{\text{det}}$ denotes the standard \yolovnine detection loss and $\mathcal{L}_{\text{lb}}$ is the load balancing term weighted by $\lambda_{\text{lb}}$.
The load balancing loss encourages uniform expert selection and is computed as:
\begin{equation}
    \mathcal{L}_{\text{lb}} = \frac{1}{I} \sum_{i=1}^{I} E \cdot \sum_{e=1}^{E} f_{i,e} \cdot P_{i,e},
\end{equation}
where $f_{i,e}$ denotes the fraction of samples routed to expert $e$ at resolution $i$, and $P_{i,e}$ represents its corresponding mean routing probability.

\section{Experiments and results} \label{sec:Experiments_and_results}

We conducted experiments on the COCO~\cite{cocoDataset} and VisDrone~\cite{Zhu2020DetectionAT} datasets, which represent everyday scenes and aerial imagery, respectively, focusing on the \textit{Person}, \textit{Vehicle}, \textit{Bicycle}, and \textit{Motorcycle} classes (the classes shared by both datasets).
The models were trained for $50$ epochs using the default hyperparameters of the \yolovnine framework~\cite{wang2024yolov9}, and our architecture employed two experts based on \yolovnine-T, with $\lambda_{\text{lb}} = 0.5$.
As in~\cite{negroni2025leveraging}, the experts were initialized from weights pretrained on their datasets (COCO or VisDrone), enabling dataset-specific specialization, while the routing modules were trained from scratch.
The best model was selected according to the \textit{mean Average Precision} (mAP) metric. 
The evaluation was conducted using \textit{mAP} and \textit{Average Recall} (AR) metrics. 
As shown in~\autoref{tab:results_performance}, our model achieved better performance compared to the other \yolovnine-T models on both datasets. 

\begin{table}[t]
    \centering
      \caption{Performance comparison between our proposed model and \yolovnine-T variants on the COCO and VisDrone (Vis.) datasets, evaluated on the classes \textit{Person}, \textit{Vehicle}, \textit{Bicycle}, and \textit{Motorcycle}}
    
    \begin{tabular}{|l|l|l|c|c|}
        \hline
        \textbf{Test Set} & \textbf{Train Set} & \textbf{Model} & \textbf{mAP@0.5:0.95} & \textbf{AR} \\ 
        \hline
        \multirow{4}{*}{COCO} 
        & COCO & \yolovnine-T  & 34.5 & 46.7 \\ 
        & COCO + Vis. & \yolovnine-T & 34.1 & 49.2 \\ 
        & COCO + Vis. & \textbf{\moe-T (Ours)} & \textbf{37.5} & \textbf{50.0} \\ 
        
        \hline
        \multirow{4}{*}{Vis.} 
        & Vis. & \yolovnine-T & 18.3 & 34.7 \\ 
        & COCO + Vis. & \yolovnine-T & 15.5 & 30.3 \\ 
        & COCO + Vis. & \textbf{\moe-T (Ours)} & \textbf{20.0} & \textbf{36.6} \\ 
        
        \hline
    \end{tabular}
    \label{tab:results_performance}
\end{table} 
The improvements appear both when each dataset is trained separately and when using a single model trained on the combined sets, showing consistent gains in both mAP and AR. These results demonstrate the effectiveness of incorporating the \moe mechanism in the feature map space by allowing experts to specialize in different visual features so that the router dynamically selects the most relevant expert for each image region.

\section{Conclusions and future work}
\label{sec:Conclusions_and_future_work}
This paper presents a hybrid object detection framework that combines the \moe paradigm within the \yolovnine architecture. By introducing adaptive routing between multiple \yolovnine-T experts, the proposed approach enables dynamic specialization in feature map space and improves detection performance and robustness compared to a single model. Experimental results on the COCO and VisDrone datasets demonstrate the potential of this design.
For future work, we plan to extend the study to additional YOLO variants of different sizes (e.g., \yolovnine-L), explore more efficient routing mechanisms, and adapt the framework to temporal data for video object detection and multi-modal inputs.
\bibliographystyle{IEEEtran}
\bibliography{references}

\end{document}